# Robust Pixel-wise Digital Image Correlation under Large Motions and Illumination Variations via a Global-to-Local Hybrid Neural Field

Guo Xianglong[1,*], Li Jianyang[1]

1. Institute of Solid Mechanics, School of Aeronautic Science and Engineering, Beihang University, Beijing 100191, China

*Corresponding author: 21051028@buaa.edu.cn

**Abstract:** Digital image correlation (DIC) is a widely used non-contact technique for full-field displacement measurement. Data-driven deep-learning DIC offers rapid inference but may generalize poorly to unseen speckle patterns, deformation modes, noise levels, and illumination conditions. Classical optimization-based DIC and recent image-constrained neural formulations avoid labeled displacement data, yet their convergence and accuracy remain vulnerable to initialization, large rigid-body motion, fine non-uniform deformation, and gray-value disturbances. This paper presents a global-to-local hybrid neural DIC framework for robust pixelwise measurement. Stage 1 estimates a dominant in-plane rigid transformation through coarse rotation search, phase correlation, and masked refinement, providing a reliable initialization under large translation and rotation. Stage 2 represents the residual displacement with a hybrid Kolmogorov-Arnold network (KAN) and multilayer perceptron (MLP), and optimizes it using a valid-overlap, illumination-robust structural loss. The local gray-value model explicitly accommodates multiplicative and additive intensity changes. Simulated and real experiments demonstrate accurate and stable displacement recovery under noise, non-uniform deformation, large rotation, illumination variation, and degraded imaging conditions.
**Keywords:** Digital image correlation; Neural displacement field; Global-to-local optimization; Illumination robustness; Large rigid-body motion.

## 1. Introduction

Digital image correlation (DIC) is a widely used non-contact optical technique for full-field displacement and strain measurement in experimental mechanics, materials characterization, and structural health monitoring [1–3]. Given reference and deformed images, DIC estimates subpixel displacement by tracking natural or artificial speckle patterns. Its basic observation is that corresponding material points retain a predictable gray-value relationship, possibly after a multiplicative or additive intensity change. At each pixel, however, gray-value consistency supplies only one scalar constraint for two displacement components. Pixelwise DIC is therefore intrinsically underdetermined and requires a spatial displacement model or regularization; the formal mapping used in this work is introduced below.

As shown in Fig. 1, the reference image $I_{ref}$ transforms into the deformed image $I_{def}$ after the specimen undergoes an in-plane displacement field. In this work, the spatial coordinate is denoted as $X = (x, y)^T$ where $x$ and $y$ represent the horizontal and vertical image coordinates, respectively. The displacement vector at $X$ is denoted as $\mathbf{d}(X)$, with its two components $u(X)$and $v(X)$ representing the horizontal and vertical displacements:

$$\mathbf{d}(X) = [u(X), v(X)]^T \tag{1}$$

Equation (1) defines the unknown displacement field to be solved in DIC. After deformation, the material point originally located at $X$ moves to a new position $X'$, which can be expressed as

$$X' = X + \mathbf{d}(X) \tag{2}$$

Equation (2) describes the geometric relationship between the reference and deformed configurations. Based on the photometric consistency assumption, the grayscale intensity of the same material point remains unchanged before and after deformation [4], [5]. Therefore, the grayscale values of the two images satisfy:

$$I_{ref}(X) \approx \alpha I_{def}\left(X + \mathbf{d}(X)\right) + \beta \tag{3}$$

Equation (3) provides the photometric governing relation of DIC under an affine grayscale model. It implies that, when the correct displacement field $\mathbf{d}(X)$ is obtained, each material point in the reference image can be mapped to its corresponding position in the deformed image, with possible grayscale variations described by $\alpha$ and $\beta$. Therefore, the fundamental task of DIC is to estimate the displacement field from the reference and deformed image pair [1], [6].

Directly solving Eq. (3) at the pixel level is underdetermined because two displacement components must be inferred from one scalar photometric constraint. Conventional subset-based DIC makes this problem tractable by representing the displacement field within each subset using a low-dimensional shape function [7]. In contrast, the present work adopts a learnable coordinate-based representation to estimate the full-field displacement, as introduced in Section 2.1.1.

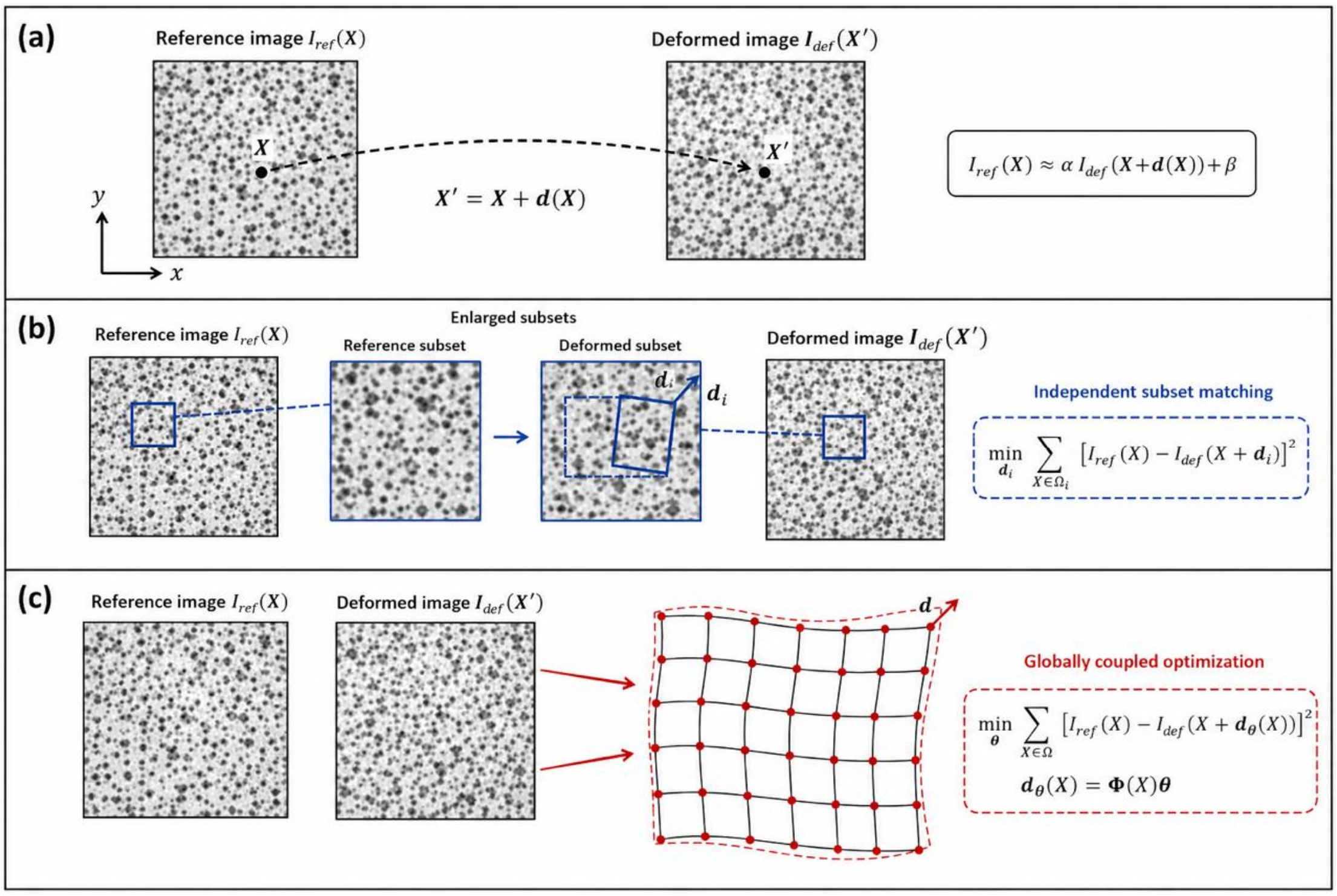

**Fig. 1. DIC image mapping and conventional optimization strategies. (a) Material-point mapping with a linear gray-value model. (b) Subset-based local DIC. (c) Finite-element-based global DIC.**

Subset-based local DIC and finite-element-based global DIC are the two canonical optimization formulations, as illustrated in Fig. 1(b) and (c). Local DIC divides the region of interest into overlapping subsets and represents the displacement within each subset by a low-order shape function. Global DIC instead uses nodal degrees of freedom and finite-element interpolation to obtain a spatially continuous displacement field over the complete region [6, 7]. Both formulations are widely accepted and continuously refined, and methods such as Smart-DIC further reduce user intervention through adaptive parameter selection [8]. Their limitations, however, are distinct. In local DIC, subset size and step size create an inherent trade-off between spatial resolution, noise resistance, and computational cost; complex or highly localized deformation may violate the assumed subset shape function, while large motion and rotation demand reliable initialization and repeated nonlinear iterations. Global DIC improves field continuity and noise resistance but depends on mesh density and interpolation order: a coarse mesh can oversmooth local deformation, whereas a fine mesh increases the size and conditioning difficulty of the coupled optimization problem. Both approaches also remain sensitive to speckle decorrelation, interpolation bias, illumination changes, and local minima. These limitations motivate dense methods that reduce manual parameter dependence while retaining optimization-based physical consistency.

The demand for dense, automated, and computationally efficient measurement has therefore stimulated data-driven deep-learning DIC. Most supervised methods adopt optical-flow or encoder-decoder architectures and learn an end-to-end mapping from a reference-deformed image pair to displacement or strain fields [9–18]. Once trained, they can deliver rapid pixelwise predictions and can handle deformation patterns that are difficult to describe with a fixed low-order subset model. Their accuracy and generalization, however, depend strongly on the diversity and fidelity of labeled training data. Because high-accuracy ground-truth fields are difficult to obtain experimentally, training commonly relies on simulated speckles and synthetic deformation. The resulting domain gap can degrade performance on unseen speckle qualities, camera noise, illumination changes, image resolutions, deformation modes, or large rigid-body motion.

Complementary to supervised image-to-field predictors, optimization-oriented neural DIC represents the displacement field with a neural function and optimizes its parameters separately for each image pair by enforcing image consistency [19, 20]. Physics-informed or mechanically constrained variants may additionally impose strain-displacement relations, compatibility, or other regularization [21–23]. PINN-DIC, KAN-DIC, and PiNetDVC exemplify this image-pair-specific formulation and do not require labeled displacement fields [24–26]. Three practical limitations remain. First, direct full-field optimization from the unaligned image pair is sensitive to initialization and can fail under large translation or rotation. Second, a single fully connected network or a pure KAN can have difficulty representing both the global deformation trend and fine local residuals. Third, raw photometric matching is vulnerable to illumination and contrast changes, haze-like degradation, and invalid boundary regions created by warping. A robust formulation must therefore coordinate initialization, displacement representation, and illumination-insensitive matching.

To address these limitations, this work develops a global-to-local hybrid neural DIC framework with three coupled components. First, masked global rigid alignment compensates for large translation and rotation before dense optimization, providing a reliable initialization. Second, a hybrid KAN-MLP neural field represents the residual displacement: the KAN supplies compact nonlinear coordinate features, while the multilayer perceptron (MLP) refines them into pixelwise displacement components. Third, a valid-overlap, illumination-robust structural loss combines local zero-normalization, Sobel gradients, and multi-scale aggregation to suppress intensity variation and invalid non-overlapping regions. Together, these components improve the robustness and accuracy of pixelwise displacement measurement under challenging deformation and imaging conditions.

The remainder of this paper is organized as follows. Section 2 presents the neural displacement representation, the proposed global-to-local optimization framework, the hybrid KAN-MLP residual model, and the illumination-robust structural loss. Section 3 reports simulated and real experiments, and Section 4 concludes the paper.

## 2. Method

### 2.1. Neural Displacement Representation and Optimization

#### 2.1.1. Displacement-field parameterization

Following PINN-DIC and KAN-DIC [24, 25], the displacement field is parameterized by a neural function whose inputs are spatial coordinates. This description refers to the network parameterization rather than to a new category of DIC. Specifically, instead of assigning independent deformation parameters to each subset, a shared network $f_\theta$ maps every reference-image coordinate $X = (x, y)^{\mathrm{T}}$ to a displacement vector $\mathbf{d}_\theta(X)$:

$$\mathbf{d}_\theta(\mathbf{X}) = f_\theta(X) \tag{4}$$

where $\theta$ denotes the trainable network parameters.

Equation (4) is a learnable alternative to the prescribed shape functions of classical DIC. In PINN-DIC [24], $f_\theta$ is a fully connected neural network, equivalently a multilayer perceptron (MLP) composed of affine layers and nonlinear activations. The same parameters are shared across the complete region of interest, so the network defines a continuous coordinate-to-displacement field rather than an image-to-field predictor trained on a population of labeled samples.

KAN-DIC [25] replaces the fixed node activations of a conventional MLP with learnable univariate functions on network edges, following the Kolmogorov-Arnold network (KAN) formulation [27]. The distinction is important: classical local or global DIC fixes the spatial basis functions and optimizes only their coefficients, whereas a neural field learns a shared nonlinear basis from the image pair and uses that basis to determine the displacement at every coordinate. As visualized in Fig. 3, the proposed model uses KAN as a compact nonlinear feature generator and an MLP head to fuse those features and recover fine residual displacement components. This division of roles motivates the hybrid architecture used in Stage 2.

Based on the predicted displacement field, the reference image can be warped to reconstruct the deformed image. The network parameters are then optimized by enforcing the photometric consistency between the reconstructed deformed image and the actual deformed image:

$$L_{photo} = \sum_{X \in \Omega} \left[ I_{ref}\left(X - D_\theta(X)\right) - I_{def}(X) \right]^2 \tag{5}$$

where $\Omega$ denotes the region of interest.

Equation (5) defines an image-pair-specific, label-free objective: coordinates are the network inputs, displacement is the output, and image consistency supplies the observation constraint. Related representations have been used in PINN-DIC, KAN-DIC, and physics-informed digital volume correlation [24–26], supporting the use of continuous neural functions for two- and three-dimensional deformation measurement.

Direct optimization of Eq. (5) nevertheless remains difficult. Large translation or rotation weakens the initial correspondence and can steer the optimizer toward an incorrect local minimum; a single neural representation may not capture both the global trend and fine residual deformation; and even the linear gray-value model cannot fully describe spatially varying illumination, local contrast changes, haze-like degradation, or invalid warped boundaries. These observations motivate a two-stage framework in which global rigid alignment is followed by hybrid KAN-MLP residual-field optimization with illumination-robust structural matching.

### 2.1.2. Overview of the proposed global-to-local framework

The framework contains two sequential estimation stages. Stage 1 estimates and compensates for the dominant global rigid-body motion. Stage 2 represents and optimizes the remaining non-rigid displacement with a hybrid KAN-MLP neural field. The illumination-robust structural loss is the Stage-2 objective, not a third estimation stage. As shown in Figs. 2 and 3, this decomposition replaces direct full-field optimization with global rigid alignment followed by dense residual estimation.

In the first stage, a masked global alignment warm-up is performed to estimate the dominant rotation and translation between the reference image $I_{ref}$ and the deformed image $I_{def}$. As shown in Fig. 2, a coarse angle search is first conducted over a predefined rotation range. For each candidate angle, phase correlation is used to estimate the corresponding translational displacement. The candidate transformations are then evaluated using a masked image-matching loss, and the best coarse transformation is further refined through iterative optimization. The output of this stage is a roughly aligned reference image, denoted as $I_{ref}^{a}$. This stage reduces the difficulty of the subsequent dense displacement estimation, especially when the image pair contains large translation, rotation, or global rigid-body motion.

In the second stage, the residual displacement field is represented by a coordinate-based hybrid KAN-MLP neural field. As shown in Fig. 3, the image coordinate $(x, y)$ is taken as the network input. The KAN module first maps the coordinate input into a nonlinear feature representation, while the MLP head further converts the learned features into a two-dimensional residual displacement vector. The residual displacement field can be expressed as

$$h(X) = f_{KAN}(X), D_{res}(X) = f_{MLP}(h(X)) \tag{6}$$

Here, $D_{res}$ denotes the residual deformation after global alignment. Compared with pure FCNN-based PINN-DIC or pure KAN-based KAN-DIC, the proposed hybrid KAN-MLP design aims to combine the compact nonlinear representation ability of KAN with the flexible local refinement capability of the MLP head. The aligned reference image $I_{ref}^{a}$ is then warped according to $D_{res}$, and the network parameters are optimized by minimizing the discrepancy between the warped reference image and the deformed image.

To improve robustness under practical imaging conditions, the proposed framework does not rely only on direct pixel-wise photometric error. Instead, a local zero-normalized Sobel structural loss is used during dense residual optimization. This loss compares local normalized gradient structures rather than raw grayscale values, making the optimization less sensitive to illumination variation, background haze, and local contrast changes. In addition, valid-overlap masking is used to suppress invalid boundary regions introduced by image warping. The final displacement field is obtained by combining the global displacement estimated in Stage 1 and the dense residual displacement estimated in Stage 2.

### 2.2. Stage 1: Masked global rigid alignment

The first stage estimates and compensates for the dominant global rigid motion before residual deformation estimation. The estimated rigid component is retained and is later composed with the residual mapping to obtain the complete reference-to-deformed mapping. Therefore, a masked global rigid alignment stage is introduced as a warm-up step.

As shown in Fig. 2, the global transformation is parameterized by a rotation angle and two translation components. Let $\alpha_c = (\theta_c, t_{xc}, t_{yc})$ denote the global rigid parameters, where $\theta$ is the in-plane rotation angle and $t_{xc}, t_{yc}$ are the translations along the horizontal and vertical directions. The transformed reference image can be written as

$$I_{ref}^{g} = W_g(I_{ref}, \alpha_c) \tag{7}$$

where $W_g(\cdot)$ denotes the differentiable global warping operation. Equation (7) means that the reference image is first globally rotated and translated to roughly match the deformed image.

To obtain a reliable initialization, a coarse-to-fine search strategy is adopted. The reference image is first rotated over a predefined angle range, and phase correlation is used to estimate the corresponding translational shift for each candidate angle. These candidate rigid transformations are evaluated by an image-matching score, and the best one is selected as the initial global alignment. This initialization reduces the risk of the optimizer being trapped near the identity transformation when large rotation or translation exists.

After coarse initialization, the global rigid parameters are further refined by minimizing a masked alignment loss. Since large translation or rotation may introduce non-overlapping regions after warping, directly computing the loss over the whole image would include invalid zero-padding areas and pixels without true correspondences. To avoid this issue, a bidirectional valid-overlap mask is constructed using a warped-ones strategy. Specifically, forward and inverse warped-ones masks are used to identify valid regions from the transformed reference image and the deformed image, respectively. Their intersection is used as the final mask, so that only pixels with valid correspondences in both images contribute to the alignment loss.

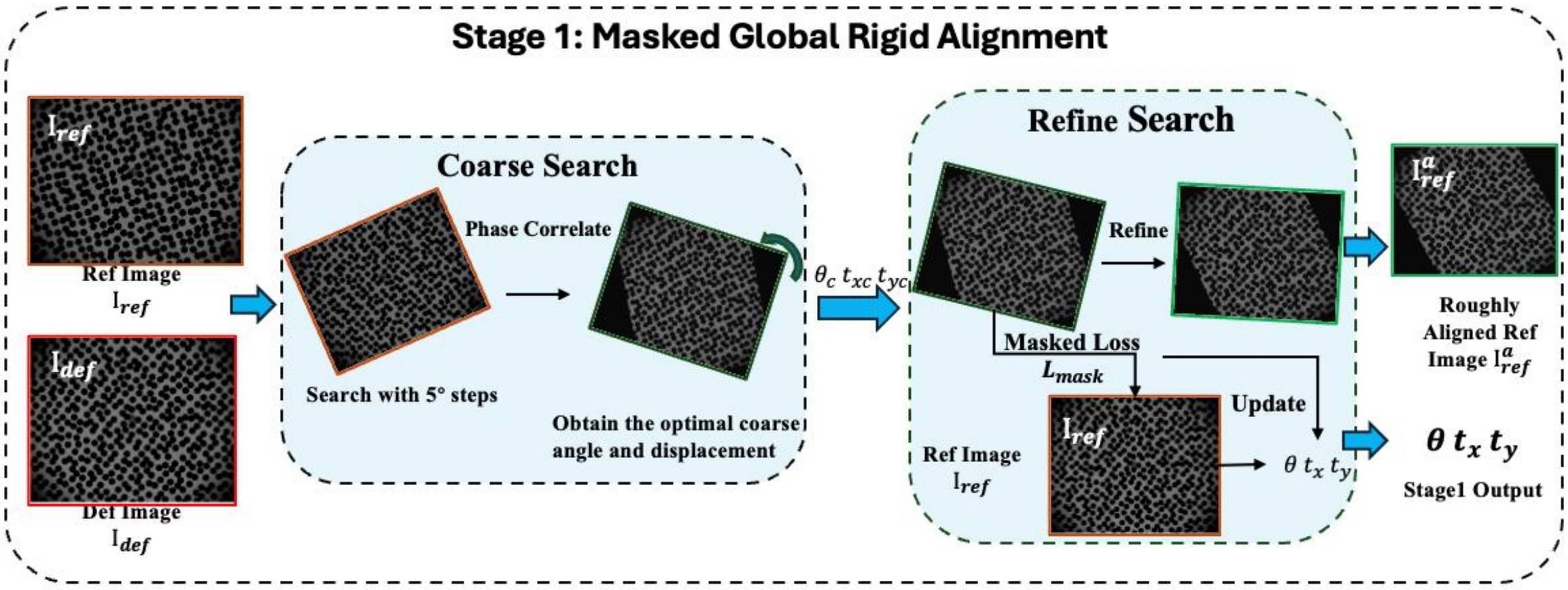

**Fig. 2. Overview of Stage 1 masked global rigid alignment. A coarse angle search and phase correlation provide the initial rotation and translation, followed by masked loss-based refinement to obtain a roughly aligned reference image.**

The masked alignment loss is then defined as a combination of a masked structural term and a masked photometric term:

$$L_{align} = L_{ZN-sobel}^{mask} + 0.2L_{photo}^{mask} \tag{8}$$

Here, $L_{ZN-sobel}^{mask}$ denotes the local zero-normalized Sobel structural loss computed within the valid-overlap region, and $L_{photo}^{mask}$ denotes the masked photometric loss. The structural term uses the local zero-normalized Sobel formulation defined in Section 2.4, whereas the complete Stage-1 objective additionally contains the masked photometric term and the overlap penalty. The photometric term is further equipped with an overlap penalty to discourage degenerate transformations that produce an excessively small valid region:

$$L_{photo}^{mask} = \frac{\sum_{X\in\Omega} M(X)\left[I_{ref}^{g}(X) - I_{def}(X)\right]^2}{\sum_{X\in\Omega} M(X) + \epsilon} + \beta[\max(0, \tau - \bar{M})]^2 \tag{9}$$

In Eq. (9), $I_{ref}^{g}$ denotes the globally warped reference image, $I_{def}$ denotes the deformed image, and $M(X)$ is the bidirectional valid-overlap mask. Here, $\varepsilon$ is a small constant introduced to prevent division by zero when the valid-overlap area is extremely small. The first term computes the mean squared gray-value difference only over pixels with valid correspondences in both images. The second term is an overlap penalty, where

$$\bar{M} = \frac{1}{|\Omega|}\sum_{X\in\Omega} M(X) \tag{10}$$

denotes the valid overlap ratio, $\tau$ is the minimum expected overlap ratio, and $\beta$ controls the penalty strength. This penalty prevents the optimizer from selecting a transformation that artificially reduces the valid matching area instead of improving image alignment.

By minimizing Eq. (8), the refined global rigid parameters are obtained as

$$\alpha_c^* = \arg min_{\alpha_c} L_{align} \tag{11}$$

The aligned reference image is then generated by applying the optimized global transformation to the original reference image:

$$I_{ref}^{a} = W_g\left(I_{ref}, \alpha_c^*\right) \tag{12}$$

### 2.3. Stage 2: Hybrid KAN-MLP residual deformation learning

After the dominant global rigid motion is compensated in Stage 1, the remaining difference between the aligned reference image $I_{ref}^{a}$ and the deformed image $I_{\mathrm{def}}$ is mainly caused by local non-rigid deformation. Therefore, Stage 2 focuses on estimating a dense residual displacement field rather than relearning the entire deformation from the original image pair.

The residual displacement field is represented by a coordinate-based hybrid KAN-MLP neural field. Given an image coordinate $X = (x, y)^T$ the network predicts the residual displacement vector as

$$\begin{bmatrix} \xi_{res}(X) \\ \eta_{res}(X) \end{bmatrix} = f_{MLP}(f_{KAN}(X; \theta_K); \theta_M) \tag{13}$$

Here, $D_{res}(X) = [\xi_{res}(X), \eta_{res}(X)]^T$ denotes the residual displacement after global alignment. The KAN module first maps the coordinate input into a nonlinear feature representation, while the MLP head further converts the learned features into a two-dimensional displacement vector. In our implementation, the MLP head is a fully connected feedforward network composed of linear layers and Tanh activation functions.

Compared with pure FCNN-based PINN-DIC or pure KAN-based KAN-DIC, the proposed hybrid representation aims to combine the compact nonlinear modeling ability of KAN with the flexible local refinement capability of the MLP head. The final layer of the MLP head is initialized close to zero, so the residual displacement starts from a near-zero field after global alignment. This design is consistent with the purpose of Stage 2: the large rigid-body motion has already been removed in Stage 1, and the neural field only needs to learn the remaining local deformation.

The predicted residual displacement field is then used to warp the aligned reference image $I_{ref}^{a}$, producing the predicted deformed image $I_{pre}$. The network parameters are optimized by minimizing the structural matching loss between $I_{pre}$ and $I_{\mathrm{def}}$, which will be introduced in Section 2.4. After optimization, the final displacement field is obtained by combining the global rigid displacement from Stage 1 and the residual displacement from Stage 2. In this way, the proposed method separates large-scale global motion from fine local deformation, improving the stability of dense displacement estimation.

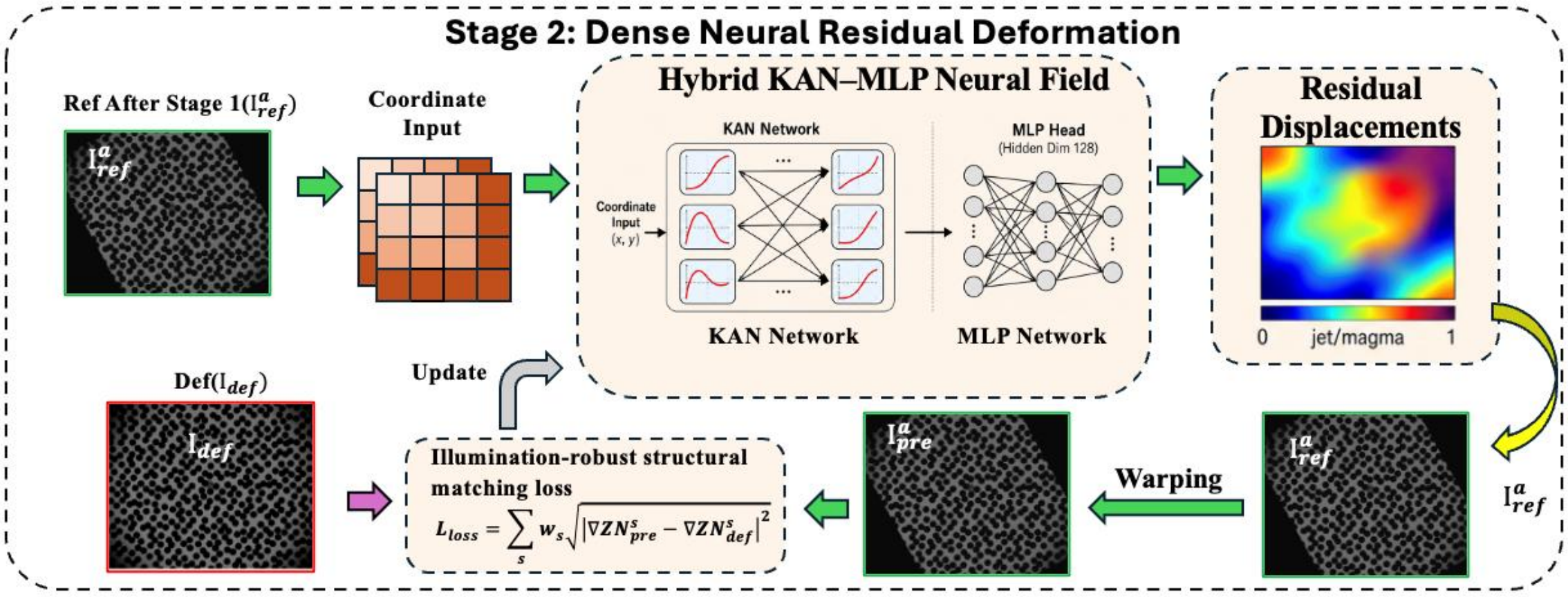

**Fig. 3. Overview of Stage 2 dense neural residual deformation estimation. A hybrid KAN-MLP neural field predicts the residual displacement and is optimized using the masked illumination-robust structural loss.**

## 2.4. Illumination-robust structural matching loss

In Stage 2, the hybrid KAN–MLP neural field is optimized by comparing the predicted deformed image with the actual deformed image. A straightforward choice is to use a pixel-wise photometric loss. However, raw gray-value matching is sensitive to practical imaging disturbances, such as non-uniform illumination, brightness drift, local contrast variation, haze-like background changes, and invalid boundary regions caused by image warping. These factors may violate the ideal brightness constancy assumption and lead to unstable optimization. Therefore, we introduce a local zero-normalized Sobel structural loss to improve the robustness of residual deformation learning [28], [29].

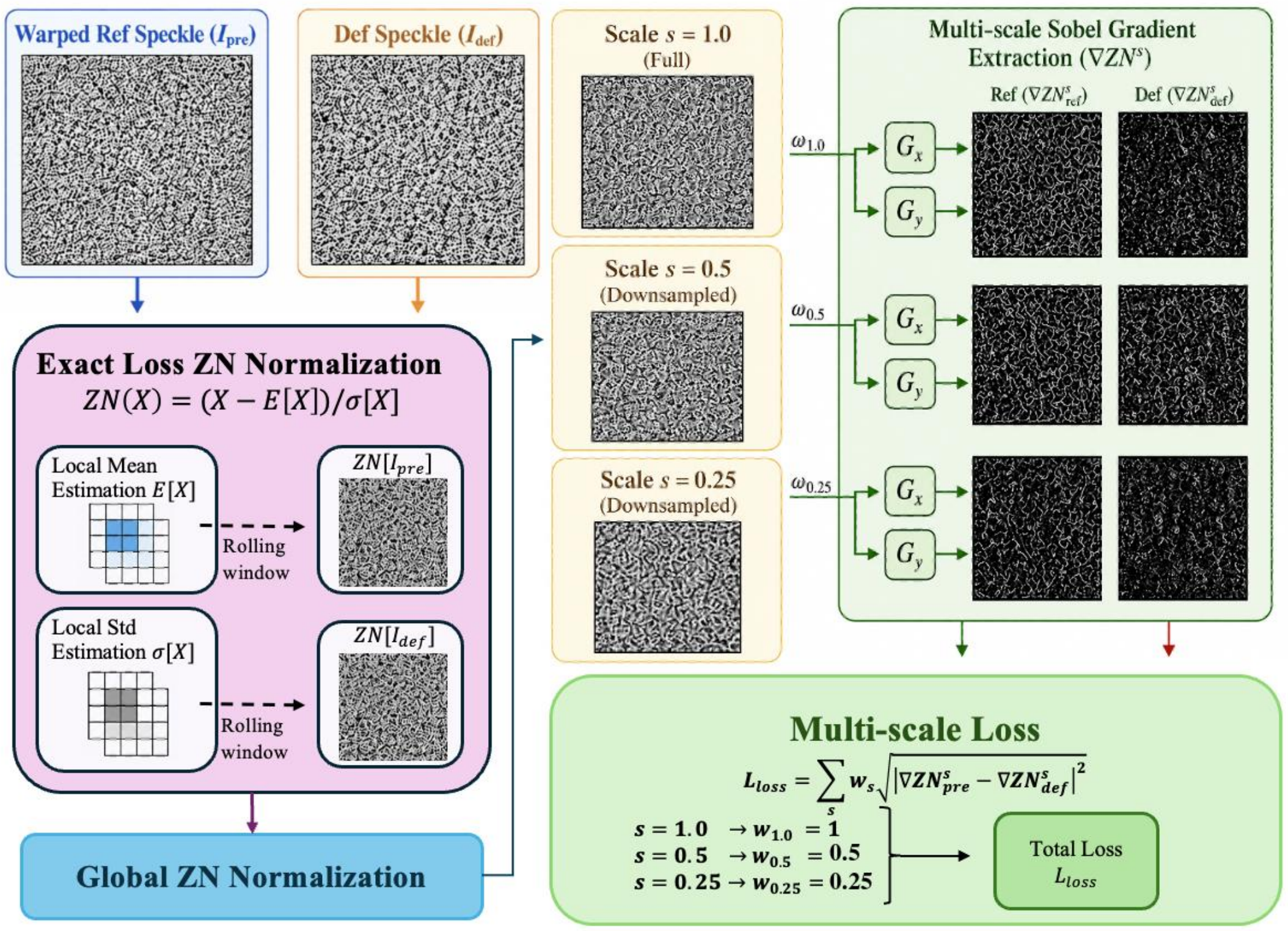

**Fig. 4. Illustration of the proposed illumination-robust structural matching loss.** The predicted deformed image $I_{pre}$ and the actual deformed image $I_{def}$ are first processed by local zero-normalization to reduce the influence of brightness drift and local contrast variation. Sobel operators are then applied to extract normalized gradient structures at multiple image scales. The final structural loss is computed as the weighted difference between the gradient fields within the valid-overlap mask, making the optimization more robust to illumination variation, haze-like background disturbance, and invalid boundary regions.

As illustrated in Fig. 4, the proposed loss consists of three main operations: local zero-normalization, Sobel gradient extraction, and multi-scale masked aggregation. Given an image $I$, local zero-normalization is first applied to reduce the influence of local brightness and contrast changes:

$$ZN(I)(X) = \frac{I(X) - \mu_N(X)}{\sigma_N(X) + \epsilon} \tag{14}$$

where $\mu_N(X)$ and $\sigma_N(X)$ denote the local mean and standard deviation computed within a neighborhood $N$ around $X$, and $\epsilon$ is a small constant for numerical stability. This normalization suppresses low-frequency intensity variations while preserving local speckle structures.

After local normalization, Sobel operators are applied to extract structural gradient information from both $I_{pre}$ and $I_{def}$. At image scale $s$, the masked Sobel structural loss is defined as [30], [31]

$$L_s = \frac{\sum_{x\in\Omega} M_s(X)\rho(\nabla \mathrm{ZN}^s(I_{pre})(X) - (\nabla \mathrm{ZN}^s(I_{def})(X))}{\sum_{x\in\Omega} M_s(X) + \epsilon} \tag{15}$$

where $\mathrm{ZN}^s$ denotes the zero-normalized image at scale $s$, $\nabla$ denotes the Sobel gradient operator; $M_s(X)$ is the valid-overlap mask resized to the corresponding scale. The function $\rho(\cdot)$ is a robust Charbonnier penalty used to reduce the influence of local outliers [32]. By comparing normalized gradient structures rather than raw pixel intensities, Eq. (15) makes the optimization less sensitive to illumination variation and haze-like background disturbance.

To further improve convergence and robustness, the structural loss is computed at multiple image scales [33]:

$$L_{zn-sobel}^{mask} = \sum_{s\in\{1,0.5,0.25\}} w_s L_s \tag{16}$$

where $w_s$ is the weight for scale $s$. In our implementation, the three scales are $1, 0.5, 0.25$, with weights $1, 0.5, 0.25$, respectively. The valid-overlap mask prevents non-overlapping regions, zero-padding artifacts, and pixels without reliable correspondences from contributing to the loss. Therefore, the proposed loss jointly improves robustness to illumination disturbance, local contrast variation, haze-like artifacts, and boundary invalidity.

In Stage 2, $L_{zn-sobel}^{mask}$ is used as the main structural matching objective between $I_{pre}$ and $I_{def}$. Combined with the global alignment in Stage 1 and the hybrid KAN-MLP residual representation in Stage 2, this loss enables stable and accurate dense displacement estimation under challenging imaging conditions.

## 3. Experiments

### 3.1. Experimental setup and implementation details

To evaluate the performance of the proposed method, a series of simulation experiments and real speckle image experiments were conducted. The simulation experiments were mainly used to quantitatively assess the displacement measurement accuracy under controlled deformation fields, while the real experiment was used to verify the applicability of the proposed method under practical imaging conditions with illumination variation and large rotation. The experiments are organized according to the three main design objectives of the proposed framework: translation and tensile tests evaluate basic displacement accuracy, large-rotation tests evaluate the global rigid-alignment strategy, degraded-image tests evaluate robustness to photometric disturbance, and real experiments assess practical applicability under combined rotation and illumination variation.

Part of the simulation data was selected from the open-source DIC benchmark dataset provided by Zhao et al. [34]. This benchmark provides simulated speckle image pairs with prescribed ground-truth displacement fields, covering representative DIC measurement scenarios such as noise-contaminated translation, rigid translation with different displacement amplitudes, rigid-body rotation, tensile deformation, and complex non-uniform deformation. Since the imposed displacement fields are available, the predicted displacement fields can be directly compared with the ground truth for quantitative evaluation. In this work, we selected several representative tests according to the objectives of the proposed method, including noise-contaminated sub-pixel translation, horizontal rigid translation, rigid-body rotation, and tensile deformation. To keep the figures readable, two representative noise levels and two representative translation amplitudes were visualized in the main text, while the remaining cases were used for quantitative analysis. In addition, degraded image simulations were constructed using random grid displacement fields and grid-like illumination degradation to further evaluate the robustness of the proposed method under non-uniform intensity disturbance.

The proposed method was compared with three representative DIC methods, including PINN-DIC [24], KAN-DIC [25], and subset-based DIC implemented using VIC2D software. PINN-DIC and KAN-DIC were selected because they are closely related coordinate-based neural DIC methods, while subset-based DIC was included as a classical traditional DIC baseline. All methods were evaluated on the same image pairs and the same region of interest. For each simulation case, both horizontal and vertical displacement components were analyzed, and the results were compared with the corresponding ground-truth displacement fields.

For PINN-DIC [24], the official implementation corresponding exactly to the published method was not available to us at the time of this study. We therefore reproduced PINN-DIC according to the network architecture, photometric objective, warm-up strategy, and optimization procedure described in the original paper. The reported PINN-DIC results should consequently be interpreted as the performance of our reproduction under the present datasets and implementation settings, rather than as a definitive assessment of the original method.

All experiments of the proposed method, PINN-DIC, and KAN-DIC were implemented in PyTorch and performed on a workstation equipped with an Intel Xeon E5-2696 v3 CPU and an NVIDIA Tesla V100 GPU. The subset-based DIC baseline was performed using VIC2D on a Lenovo ThinkPad T460p laptop equipped with an Intel

Core i5-6440HQ CPU. For the VIC2D baseline, different subset settings were used according to the measurement scenario. In the simulated deformation experiments, a subset size of 31 pixels and a step size of 1 pixel were used. In the real large-rotation experiments, a larger subset size of 151 pixels and a step size of 1 pixel were adopted to improve matching stability under severe rotation and illumination variation. The corresponding VIC2D setting was kept fixed within each experiment group.

For the proposed method, Stage 1 initialized the global rigid alignment using coarse rotation search and phase correlation, followed by masked loss-based refinement. The Stage-1 refinement was optimized using limited-memory Broyden–Fletcher–Goldfarb–Shanno(L-BFGS) algorithm for 200 iterations. In Stage 2, the dense residual displacement field was represented by the proposed hybrid KAN-MLP neural field and optimized using L-BFGS for 300 iterations. The KAN module used spline order 3 and grid size 16, and mapped the two-dimensional coordinate input to a 32-dimensional nonlinear feature representation. The MLP head then mapped this feature representation to a two-dimensional residual displacement vector. Specifically, the MLP head consisted of two hidden fully connected layers with 32 neurons each, Tanh activation functions, and a final output layer with two neurons corresponding to the horizontal and vertical residual displacement components. The final layer of the MLP head was initialized close to zero so that the residual displacement started from a near-zero field after global alignment. Unless otherwise specified, the same hyperparameters of the proposed method were used for all simulation and real experiments.

For quantitative evaluation, the mean absolute error (MAE) was adopted as the primary metric. To avoid boundary artifacts caused by image warping, interpolation, and incomplete overlap near the image borders, a 20-pixel margin was cropped from all sides before computing the MAE for every method. The MAE was calculated only within the remaining valid central region:

$$MAE = \frac{1}{N}\sum_{i=1}^{N}\left|d_i^{pred} - d_i^{gt}\right| \tag{17}$$

where the two terms inside the summation denote the predicted and ground-truth displacements at valid pixel i, and N is the number of evaluated pixels after boundary cropping. In addition to MAE, full-field error maps are used to show the spatial distribution of measurement error. For representative cases, displacement or error profiles are also extracted along selected cross-sections to compare local measurement accuracy.

For the real experiment, ground-truth displacement fields were not directly available. Therefore, the evaluation focused on the gray-field reconstruction consistency between the warped reference image and the actual deformed image. Following residual gray-field based matching quality assessment in DIC, the gray-level residual field (GLR) [35] was used as an auxiliary indicator to assess whether the estimated displacement field could correctly reconstruct the deformed speckle image under real rotation and illumination variation. In this work, GLR is defined as the absolute gray-value difference between the actual deformed image and the warped reference image.

### 3.2. Simulated deformation experiment

#### 3.2.1. Noise-contaminated subpixel translation

Noise-contaminated sub-pixel translation experiments were first conducted to evaluate the robustness of different methods under gray-value disturbance. In this test, the imposed deformation is a horizontal translation of 0.3 pixels, while the ground-truth vertical displacement remains zero. Two representative noisy image pairs were selected from the benchmark dataset, corresponding to a relatively low-noise case and a high-noise case. The proposed method, VIC2D, PINN-DIC, and KAN-DIC were applied to the same image pairs. Since the ground-truth displacement field is available, the predicted displacement fields were directly compared with the imposed deformation. To avoid boundary artifacts caused by image warping and interpolation, a 20-pixel margin was cropped from all sides before computing the MAE.

Figure 5 presents the representative results of the noise-contaminated sub-pixel translation experiments. In the low-noise case, the proposed method reconstructs a nearly uniform horizontal displacement field and preserves a near-zero vertical displacement field, both of which are consistent with the ground truth. By contrast, VIC2D exhibits noticeable random fluctuations, indicating a higher sensitivity to image noise. Under the present test setting, our reproduced PINN-DIC implementation does not recover the prescribed displacement field and exhibits large deviations in both displacement components. KAN-DIC provides a relatively stable vertical displacement field, but its horizontal displacement estimation is clearly biased from the imposed value. In the high-noise case, the degradation of all compared methods becomes more evident. Nevertheless, the proposed method still maintains the closest overall agreement with the ground truth and yields the lowest error concentration among the four methods.

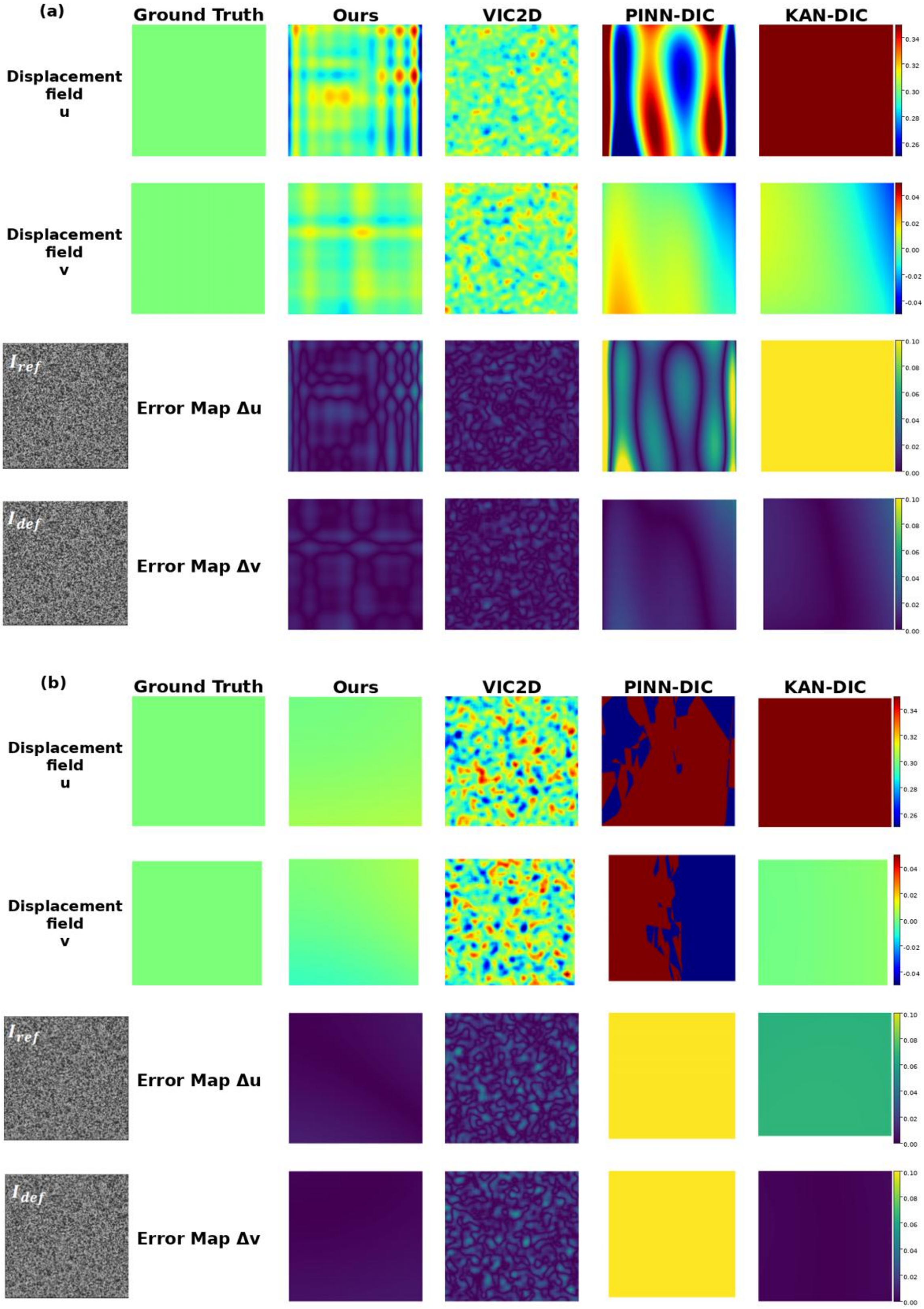

**Fig. 5. Results of noise-contaminated sub-pixel translation experiments under two representative noise levels.**(a) Low-noise case.(b) High-noise case. The imposed deformation is a horizontal translation of 0.3 pixels, with zero vertical displacement. The displacement fields $u$ and $v$, together with their corresponding error maps $\Delta u$ and $\Delta v$, are shown for the proposed method, VIC2D, PINN-DIC, and KAN-DIC. The input reference and deformed speckle

images, $I_{ref}$ and $I_{def}$ are shown in the lower-left corner of each panel. MAE values are computed after cropping a 20-pixel boundary region. The proposed method produces the closest displacement fields to the ground truth and exhibits the lowest error concentration under both noise conditions.

The quantitative MAE results are summarized in Table 1. The proposed method achieves the best or comparable accuracy under low noise and the lowest MAE under high noise, indicating improved robustness to noise and gray-value disturbance. In the low-noise case, it accurately recovers the imposed 0.3-pixel horizontal translation while maintaining a near-zero vertical displacement. As noise increases, VIC2D exhibits stronger random fluctuations, KAN-DIC produces a biased horizontal estimate, and PINN-DIC fails to converge to a physically meaningful field. The corresponding high-noise PINN-DIC entry is therefore reported as "Failed" rather than as an extreme numerical value. These results show that local zero-normalized structural matching stabilizes subpixel displacement estimation under noisy imaging conditions.

**Table 1. MAE comparison for noise-contaminated sub-pixel translation experiments.** The MAE values of the horizontal and vertical displacement fields are reported for the low-noise and high-noise cases. All MAE values are computed after cropping a 20-pixel boundary region.

| Noise Level | Method | MAE of $u$ | MAE of $v$ | MAE |
|---|---|---|---|---|
| **Low Noise** | VIC2D | **0.0063** | 0.0069 | **0.0067** |
| | PINN-DIC | 0.0378 | 0.0098 | 0.0238 |
| | KAN-DIC | 0.1669 | 0.0083 | 0.0876 |
| | **Ours** | 0.0095 | **0.0049** | 0.0072 |
| **High Noise** | VIC2D | 0.0111 | 0.0116 | 0.0114 |
| | PINN-DIC | Failed | Failed | Failed |
| | KAN-DIC | 0.0649 | 0.0008 | 0.0329 |
| | **Ours** | **0.0022** | **0.0027** | **0.0025** |

### 3.2.2. Rigid translation with different displacement amplitudes

Rigid translation experiments were conducted to evaluate the measurement accuracy of different methods under different displacement amplitudes and speckle patterns. Two representative horizontal translation amplitudes, 0.5 pixels and 1.0 pixels, were selected from the benchmark dataset [34], corresponding to sub-pixel and pixel-level rigid-body translations, respectively. To further examine the influence of speckle pattern quality, two different reference speckle images were used in this test. The proposed method, VIC2D, PINN-DIC, and KAN-DIC were applied to the same image pairs, and the predicted horizontal displacement field (u) was compared with the ground truth. As in the previous experiments, a 20-pixel boundary region was cropped before computing the MAE to avoid boundary artifacts caused by warping and interpolation.

Figure 6 presents the measured (u)-displacement fields for the rigid translation experiments. For both speckle patterns and both displacement amplitudes, the proposed method produces displacement fields that are close to the imposed uniform translation. VIC2D shows noticeable spatial fluctuations, especially under the first speckle pattern, indicating its sensitivity to local speckle quality and noise. PINN-DIC and KAN-DIC can recover the general displacement level in some cases, but their results show obvious bias or spatially non-uniform artifacts. In contrast, the proposed method maintains stable displacement estimation across different speckle patterns and displacement amplitudes, demonstrating that the hybrid KAN-MLP residual representation can preserve basic translation measurement accuracy while improving the stability of coordinate-based neural DIC.

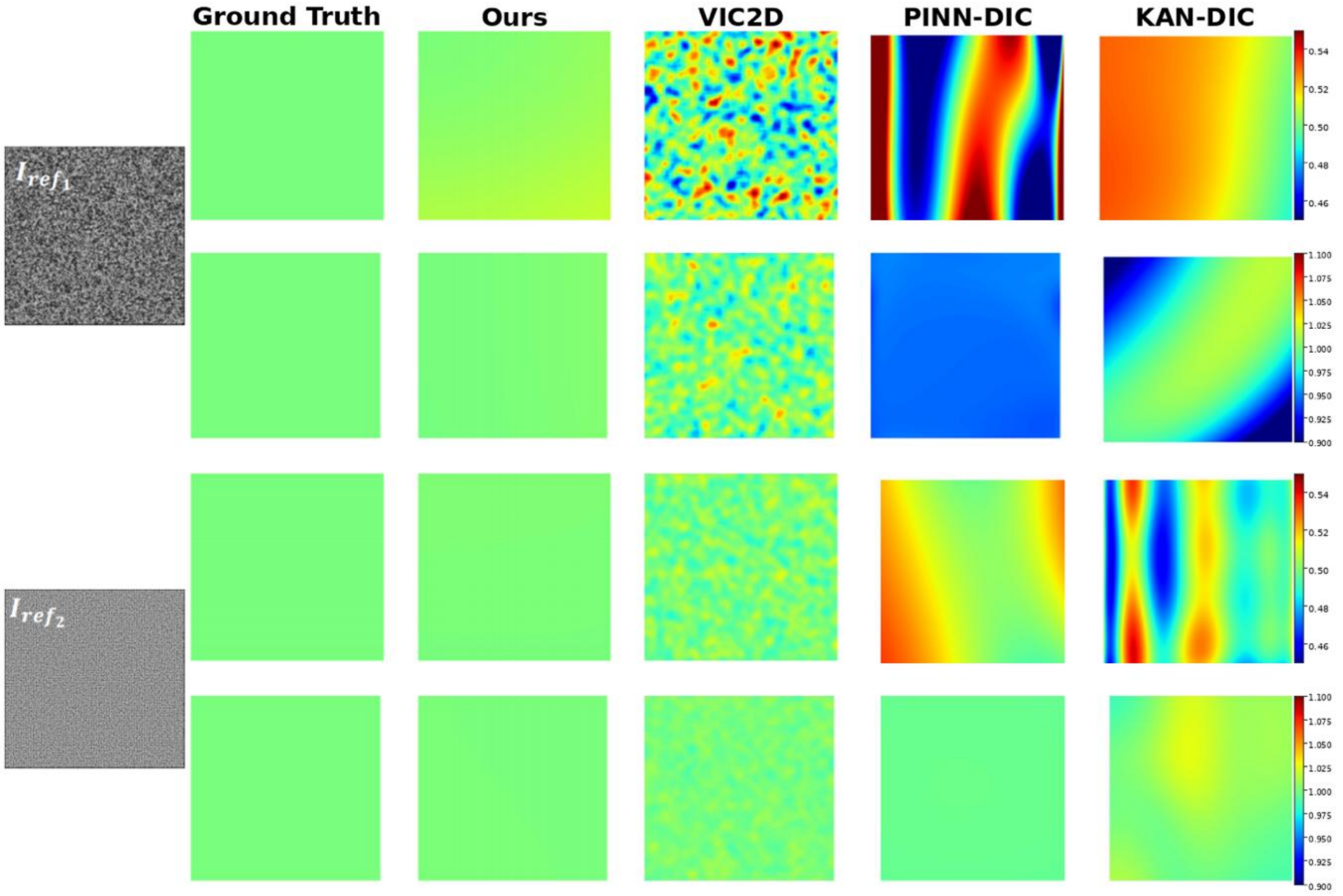

**Fig. 6. Results of rigid translation experiments with different displacement amplitudes and speckle patterns.** Two reference speckle images, $I_{ref}$ and $I_{ref_2}$, are shown in the left column. For each speckle image, two horizontal translation amplitudes are evaluated: 0.5 pixels and 1.0 pixels. The top two rows correspond to $I_{ref}$, and the bottom two rows correspond to $I_{ref_2}$. Only the horizontal displacement component (u) is shown because the imposed deformation is pure horizontal translation. The columns compare the ground truth, the proposed method, VIC2D, PINN-DIC, and KAN-DIC.

The quantitative results are summarized in Table 2. The proposed method achieves consistently low MAE values across both speckle patterns and displacement amplitudes, indicating that it can accurately recover rigid horizontal translations from sub-pixel to pixel-level motion. Compared with VIC2D, PINN-DIC, and KAN-DIC, the proposed method shows better stability across different speckle qualities. This result further confirms that the proposed hybrid KAN-MLP residual representation preserves the basic displacement measurement accuracy of DIC while reducing the spatial fluctuation observed in the compared methods.

**Table 2. MAE comparison for rigid translation experiments with different displacement amplitudes and speckle patterns.** The MAE values of the horizontal displacement component $u$ are reported for two speckle images and two imposed horizontal translations. All MAE values are computed after cropping a 20-pixel boundary region.

| Speckle Image | Translation amplitude(pixels) | VIC2D | PINN-DIC | KAN-DIC | Ours |
|---|---|---|---|---|---|
| $I_{ref}$ | 0.5 | 0.0112 | 0.0411 | 0.0198 | **0.0037** |
| $I_{ref}$ | 1.0 | 0.0111 | 0.0532 | 0.0255 | **0.0008** |
| $I_{ref_2}$ | 0.5 | 0.0030 | 0.0114 | 0.0140 | **0.0002** |
| $I_{ref_2}$ | 1.0 | 0.0028 | 0.0044 | 0.0080 | **0.0001** |

### 3.2.3. Tensile deformation

Tensile deformation experiments were conducted to evaluate the performance of different methods under regular but spatially varying displacement fields. Unlike rigid-body translation, tensile deformation introduces a smooth displacement gradient across the image, making it suitable for testing whether a DIC method can accurately recover continuous deformation variation while maintaining low measurement error. Representative tensile deformation cases were selected from the benchmark dataset [34], where the imposed displacement fields are known and can therefore be used as ground truth for quantitative comparison.

Figure 7 shows the full-field measurement results of the tensile deformation experiment. The proposed method, VIC2D, PINN-DIC, and KAN-DIC were applied to the same reference-deformed image pair. The horizontal and vertical displacement fields, together with the corresponding error maps, were compared with the ground truth. As shown in Fig. 7, the proposed method accurately reconstructs the global tensile deformation trend and yields displacement fields that are visually consistent with the imposed deformation. VIC2D exhibits noticeable local fluctuations, while KAN-DIC shows visible bias and structured error patterns. The reproduced PINN-DIC implementation did not converge to a physically meaningful displacement field under the present test settings. In contrast, the proposed method remains close to the ground truth in both the horizontal and vertical displacement components, with lower overall error concentration.

A quantitative comparison further supports this observation. After cropping a 20-pixel boundary region, the proposed method achieves $MAE_u = 0.010453$ and $MAE_v = 0.008694$, whereas VIC2D gives $MAE_u = 0.023541$ and $MAE_v = 0.010483$. This corresponds to an error reduction of approximately 55.6% in the horizontal displacement and 17.1% in the vertical displacement. These results indicate that the proposed method can more accurately recover the smooth tensile deformation field while suppressing local measurement fluctuations.

To further evaluate the displacement recovery along the deformation direction, the error profiles of the horizontal displacement $u$ extracted from the middle row are plotted in Fig. 8. Specifically, the predicted $u$-displacement profile of each method is subtracted from the corresponding ground-truth profile, so that an ideal result should remain close to zero across the entire row. As shown in Fig. 8, the proposed method yields the smallest error magnitude and the most stable error distribution, whereas VIC2D presents stronger local fluctuations and KAN-DIC shows larger systematic deviation. The PINN-DIC result is not included in the profile comparison because the corresponding full-field solution failed to converge to a physically meaningful displacement field. These results further confirm that the proposed framework can more accurately recover smooth tensile displacement gradients while maintaining stable full-field estimation.

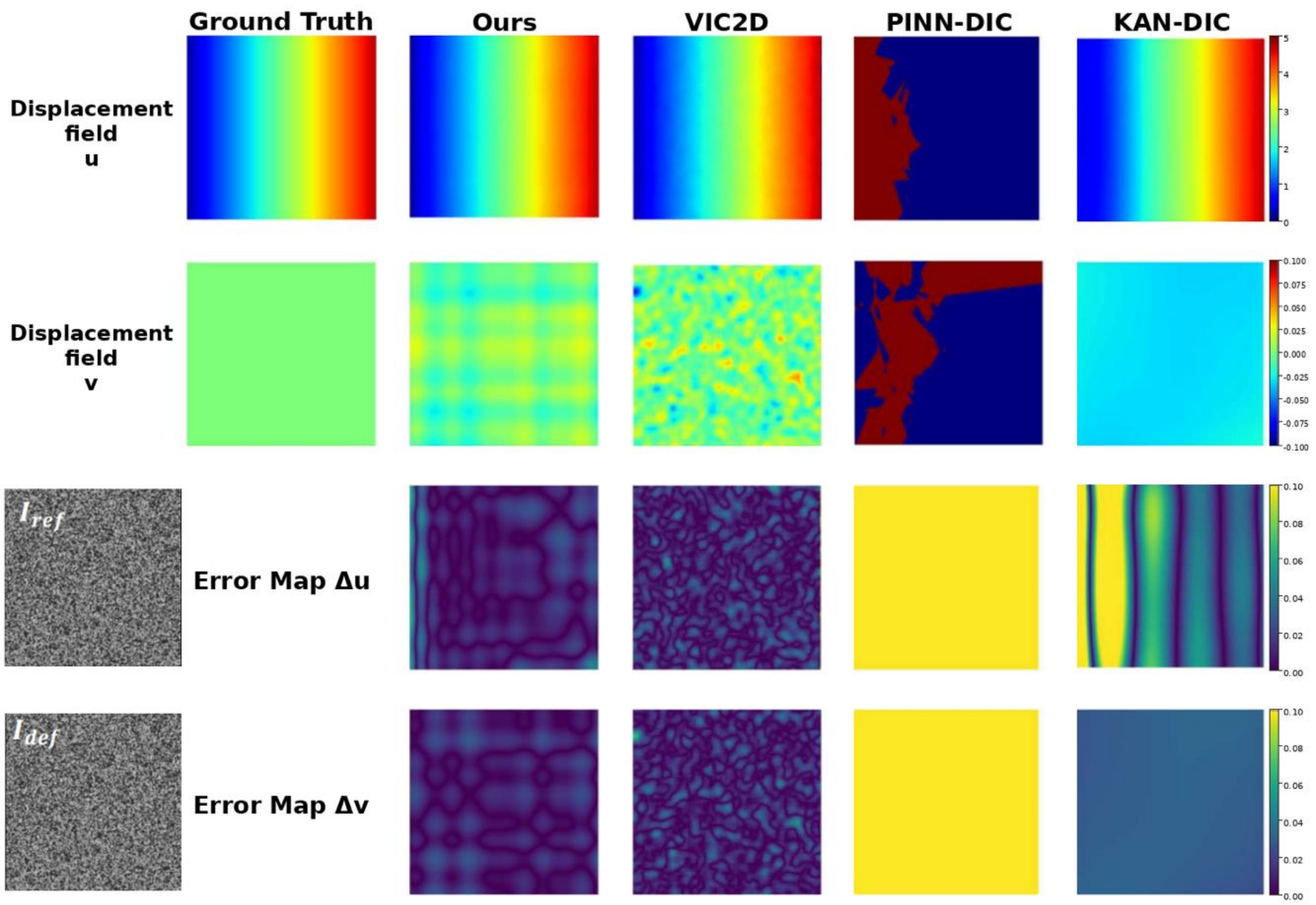

**Fig. 7. Results of tensile deformation experiments.** The horizontal displacement field $u$, vertical displacement field $v$, and the corresponding absolute error maps $\Delta u$ and $\Delta v$ are shown for the ground truth, the proposed method, VIC2D, PINN-DIC, and KAN-DIC. The reference and deformed speckle images, $I_{ref}$ and $I_{def}$, are shown in the left column. The proposed method exhibits lower full-field displacement errors. PINN-DIC fails to produce a physically meaningful displacement field in this case.

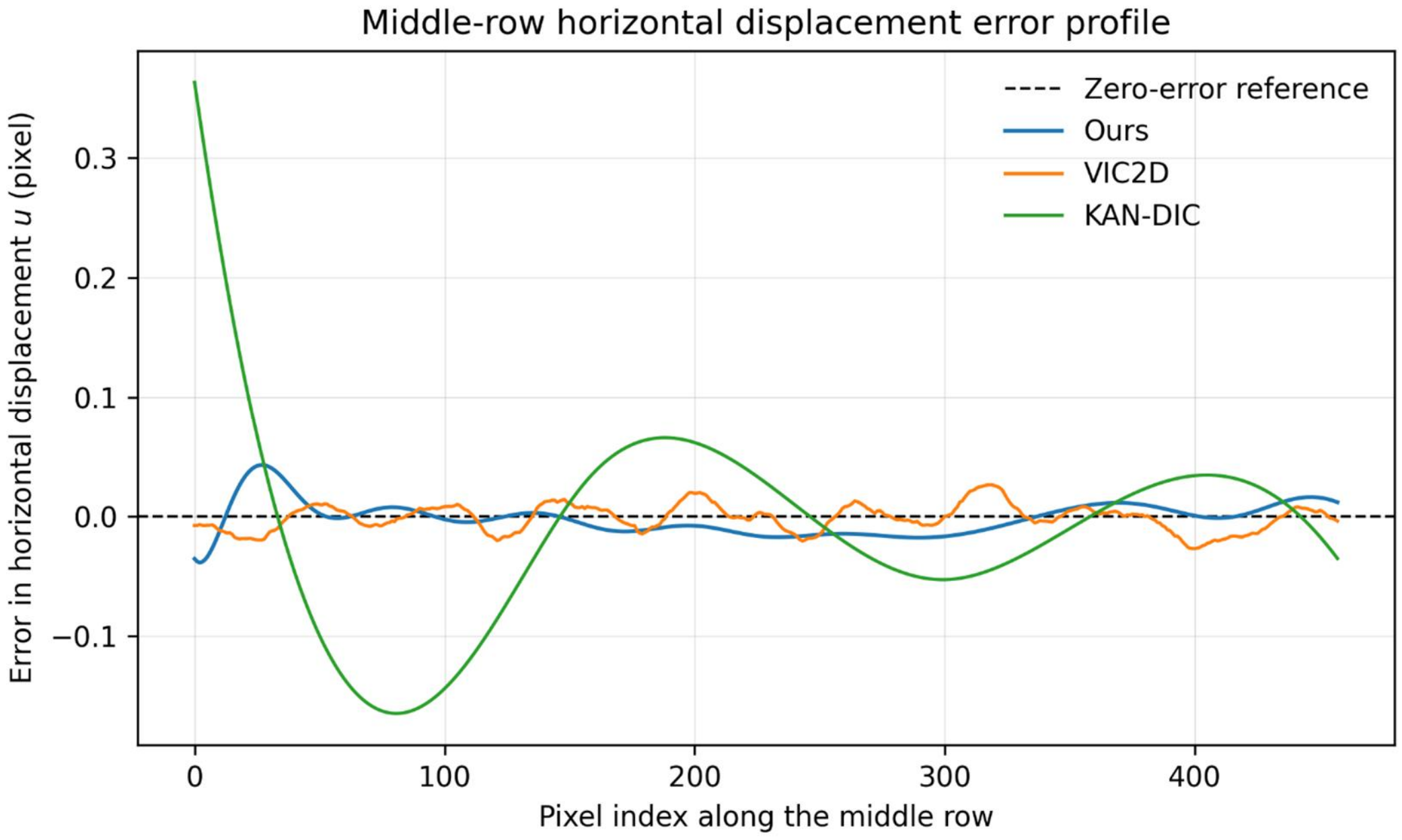

**Fig. 8. Middle-row error profiles of the horizontal displacement $u$ for the tensile deformation experiment.** For each method, the $u$-displacement values extracted from the middle row are subtracted from the corresponding ground-truth profile, and the resulting error curves are plotted from left to right. The horizontal zero line indicates

perfect agreement with the ground truth. The proposed method remains closest to zero across the row, while VIC2D shows stronger local fluctuations and KAN-DIC exhibits larger systematic deviation. The PINN-DIC result is omitted because the corresponding full-field solution failed.

### 3.3. Large rotation deformation

Large rotation deformation experiments were conducted to evaluate the robustness of the proposed method under large global rigid-body motion. Unlike pure translation or tensile deformation, rigid-body rotation introduces large spatially varying displacements across the image. For coordinate-based neural DIC methods that directly optimize the displacement field from the original image pair, such large global motion may lead to poor initial correspondence and cause the optimization to converge to an incorrect local minimum. Therefore, this experiment was designed to verify the effectiveness of the proposed masked global rigid alignment stage.

A representative 30° rotation case was selected from the benchmark dataset [34] as a challenging test of large rigid-body motion. The reference speckle image was rotated about the image center to generate the deformed image, and the corresponding ground-truth displacement field was used for evaluation. The proposed method, VIC2D, PINN-DIC, and KAN-DIC were applied to the same image pair.

Figure 9 shows the displacement measurements for the 30° rotation case. The proposed method reconstructs both displacement components, and the recovered fields closely follow the ground truth. Its results are comparable to those of VIC2D, indicating reliable recovery under large rotation. By contrast, under the present initialization and optimization settings, our reproduced PINN-DIC implementation did not converge to the prescribed rotation-induced field, while KAN-DIC showed substantial deviations from the ground-truth displacement pattern. The error maps further show low error concentration for the proposed method and VIC2D, whereas the two neural baselines without explicit global alignment fail under this challenging condition.

These results demonstrate that the proposed global-to-local strategy is effective for large rigid-body motion. By first removing the dominant rotation and translation through the masked global alignment stage and then estimating the residual deformation, the proposed method substantially improves the robustness of coordinate-based neural DIC under large rotation. Although VIC2D remains competitive in this benchmark case, the proposed method shows that a self-supervised neural DIC framework can achieve comparable performance while retaining the flexibility of coordinate-based dense deformation modeling.

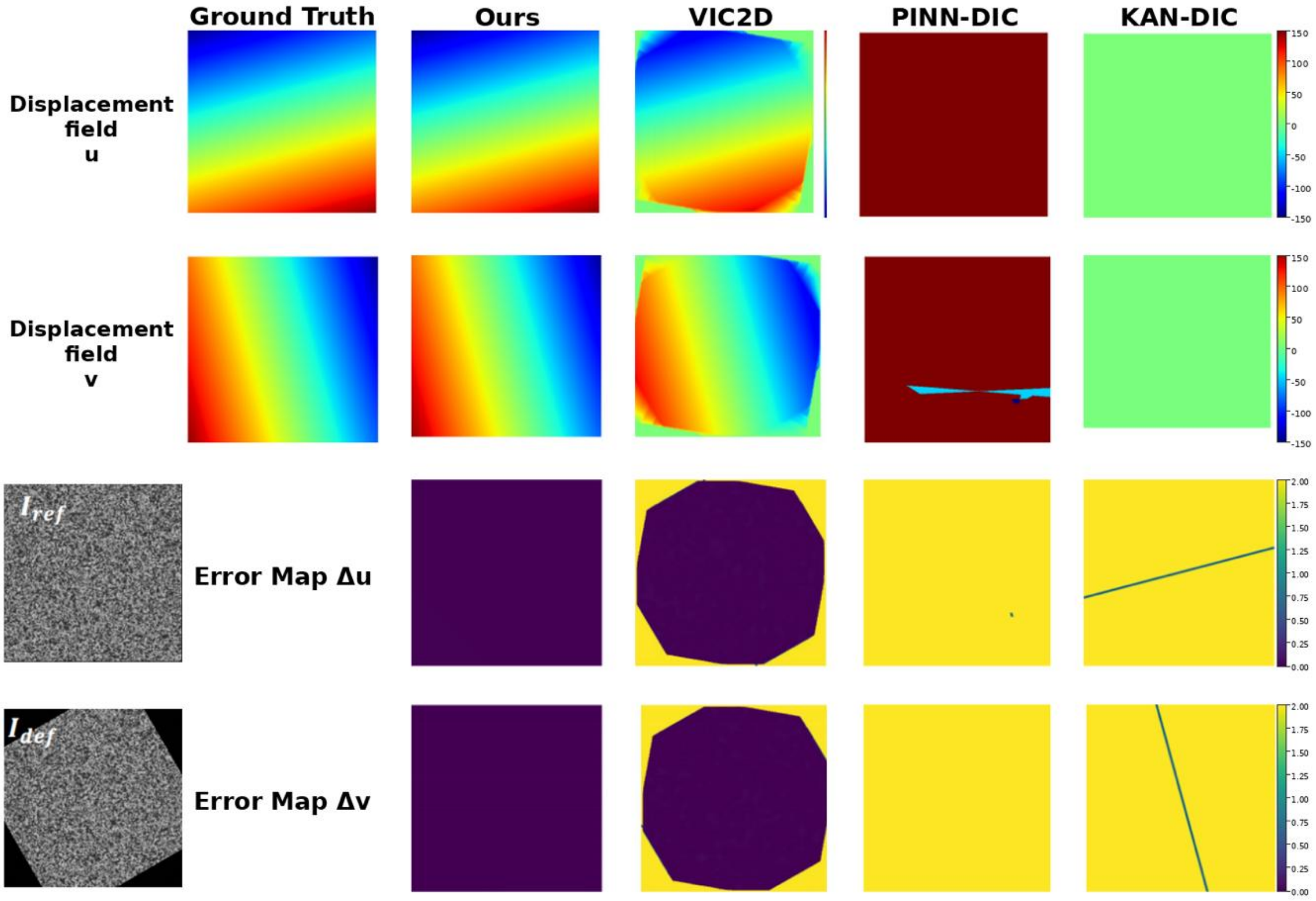

**Fig. 9. Results of a representative 30° large rotation deformation experiment.** The horizontal displacement field $u$, vertical displacement field $v$, and the corresponding absolute error maps $\Delta u$ and $\Delta v$ are shown for the ground truth, the proposed method, VIC2D, PINN-DIC, and KAN-DIC. The reference and deformed speckle images, $I_{ref}$ and $I_{def}$, are shown in the left column. The proposed method successfully recovers the large rotation-induced displacement field and achieves results comparable to VIC2D. In contrast, PINN-DIC and KAN-DIC fail to produce physically meaningful displacement fields in this case.

### 3.4. Degraded image simulation

To further evaluate the robustness of the proposed method under degraded imaging conditions, degraded image simulation experiments were conducted. This experiment aims to mimic practical optical disturbances such as non-uniform illumination, haze-like contrast attenuation, and heatwave-like image degradation. Different from the previous benchmark cases, where the image quality is relatively controlled, this experiment simultaneously introduces random deformation and spatially varying image degradation, thereby providing a more challenging test for DIC measurement.

The degraded image pairs were generated through a grid-interpolation strategy. Similar synthetic displacement generation has been used in previous deep-learning-based DIC studies, where random displacements are assigned to sparse grid nodes and then interpolated to obtain a dense and smooth displacement field [11]. Following this idea, we first generated a random speckle image as the reference image $I_{ref}$. Then, sparse control grids were defined over the image domain, and random displacement values were assigned to the grid nodes in both the horizontal and vertical directions. These sparse displacement values were interpolated to the full image resolution to obtain smooth dense displacement fields $u(x, y)$ and $v(x, y)$. The clean deformed image $I_t$ was then generated by warping the reference image according to the prescribed displacement field.

To simulate spatially varying image degradation, the same grid-interpolation strategy was used to generate the degradation parameter fields. Specifically, random values were assigned to sparse grid nodes for the atmospheric light field $A(x, y)$ and the transmission field $t(x, y)$, and interpolation was then used to obtain full-resolution smooth parameter maps. The degraded deformed image $J(x, y)$ was generated by the following haze-like image formation model:

$$J(x,y) = I_t(x,y)t(x,y) + A(x,y)[1 - t(x,y)]. \tag{18}$$

Here, $I_t(x, y)$ denotes the clean deformed speckle image, $J(x, y)$ denotes the degraded deformed image, $t(x, y)$ controls the local transmission or contrast attenuation, and $A(x, y)$ represents the spatially varying environmental light component. In this way, both the deformation field and the degradation field vary smoothly but randomly over the image domain. Although this model is simplified, it provides a controllable way to introduce non-uniform intensity disturbance and local contrast degradation into speckle images.

Two representative degraded-image cases were generated in this experiment. Each case uses a different random speckle image, random displacement field, and random degradation field. The proposed method, VIC2D, PINN-DIC, and KAN-DIC were applied to the same degraded image pairs. Since the imposed displacement fields are known, the predicted displacement fields were directly compared with the ground truth. The MAE values of both displacement components were computed after cropping a 20-pixel boundary region.

Figure 10 shows the representative degraded image simulation results. Under the combined influence of random deformation and non-uniform image degradation, the compared methods show different levels of performance degradation. VIC2D is affected by local gray-value fluctuation and produces noisier displacement fields. PINN-DIC and KAN-DIC, which rely on direct image reconstruction losses, are more sensitive to the spatially varying degradation and may fail to recover the correct displacement distribution. In contrast, the proposed method remains more stable because the local zero-normalized Sobel structural loss reduces the influence of raw intensity variation and emphasizes local speckle structure consistency.

The quantitative MAE results are summarized in Table 3. The proposed method achieves the lowest or most stable error in both degraded-image cases, demonstrating its improved robustness to non-uniform intensity degradation. These results indicate that the proposed framework can maintain reliable displacement estimation even when random deformation and spatially varying image degradation coexist.

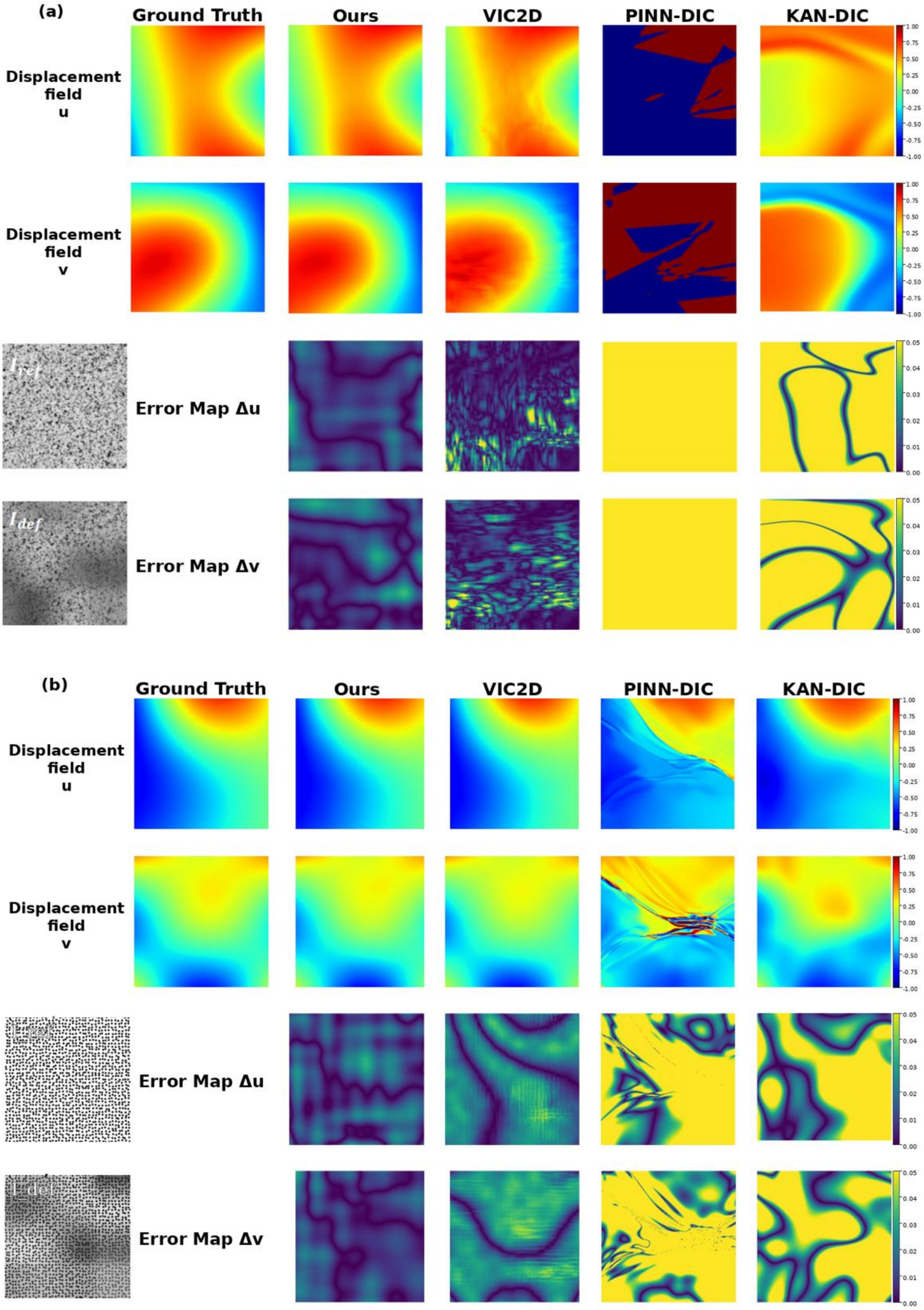

**Fig. 10. Results of degraded image simulation experiments.** (a) and (b) show two representative degraded-image cases generated with different random speckle images, random displacement fields, and grid-interpolated degradation fields. For each case, the reference image $I_{ref}$ and degraded deformed image $I_{def}$ are shown in the left column. The first two rows show the horizontal and vertical displacement fields, $u$ and $v$, and the last two rows show the corresponding absolute error maps, $\Delta u$ and $\Delta v$. The proposed method, VIC2D, PINN-DIC, and KAN-DIC are compared with the ground truth. The proposed method more accurately recovers the random displacement fields under spatially varying image degradation, while the compared methods exhibit stronger local fluctuations, structural artifacts, or failed displacement estimation. Error maps are clipped to 0.05 pixels for visualization.

**Table 3. MAE comparison for degraded image simulation experiments.** The MAE values of the horizontal and vertical displacement components are reported for the two degraded-image cases shown in Fig. 10. All MAE values are computed after cropping a 20-pixel boundary region. "Failed" indicates that the corresponding method did not converge to a physically meaningful displacement field.

| Speckle Image | Method | MAE of $u$ | MAE of $v$ | MAE |
|---|---|---|---|---|
| **(a)** | VIC2D | 0.0090 | 0.0104 | 0.0097 |
| | PINN-DIC | Failed | Failed | Failed |
| | KAN-DIC | 0.1754 | 0.1164 | 0.1459 |
| | **Ours** | **0.0094** | **0.0090** | **0.0092** |
| **(b)** | VIC2D | 0.0192 | 0.0231 | 0.0212 |
| | PINN-DIC | 0.0956 | 0.1378 | 0.1167 |
| | KAN-DIC | 0.0581 | 0.0594 | 0.0587 |
| | **Ours** | **0.0086** | **0.0114** | **0.0100** |

### 3.5. Real experimental validation

To further evaluate the practical applicability of the proposed method, real speckle image experiments were conducted under large in-plane rotation and illumination variation. Unlike the simulated experiments, the real images contain several practical disturbances, including non-uniform illumination, local contrast variation, partial disappearance and reappearance of speckle regions, and invalid boundary regions caused by image warping. Since the ground-truth displacement field is unavailable in the real experiment, the validation focuses on the reasonableness of the estimated displacement fields and the gray-field reconstruction consistency between the warped reference image and the actual deformed image. For the VIC2D baseline, a subset size of 151 pixels was used in the real large-rotation experiments to improve matching stability under severe rotation and illumination variation. The same VIC2D setting was kept fixed for all real rotation cases.

#### 3.5.1. Real rotation experiment under illumination variation

Figure 11 shows the real speckle image sequence used in this experiment. The leftmost image is the reference speckle image, and the remaining images are deformed images captured after approximately $50°$, $100°$, and $150°$ in-plane rotations, respectively. The rotation introduces large rigid-body motion and changes the visible speckle region. As a result, some speckle regions disappear from the field of view, while new regions appear after rotation. In addition, the illumination distribution varies during the experiment, leading to non-uniform gray-value changes between the reference and deformed images.

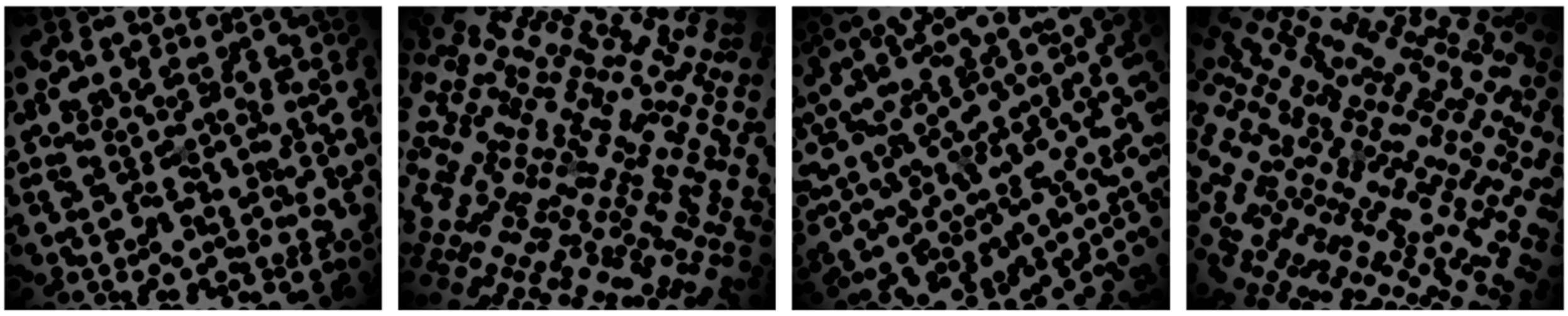

**Fig. 11. Real speckle image sequence under large rotation and illumination variation.** The leftmost image is the reference speckle image, and the remaining images are deformed images captured after approximately $50°$, $100°$, and $150°$ in-plane rotations. The real image sequence contains large rigid-body motion, non-overlapping speckle regions, boundary changes, and illumination variation.

In this real experiment, the proposed method was compared with VIC2D. PINN-DIC and KAN-DIC were not included in this comparison because the previous large-rotation simulation experiments have shown that these direct coordinate-based neural DIC methods fail to produce physically meaningful displacement fields under large global rotation. VIC2D was therefore used as the classical subset-based DIC reference method. The proposed method first performs masked global rigid alignment to compensate for the dominant rotation and translation, and then estimates the dense residual displacement field using the hybrid KAN-MLP neural representation.

### 3.5.2. Gray-field consistency evaluation using GLR

Since the ground-truth displacement fields are unavailable in the real experiment, the residual gray-field consistency was used as an indirect validation criterion. Residual gray-field or optical residual analysis is commonly used in DIC to evaluate whether the estimated displacement field can correctly map the reference image to the deformed image [35]. Following this idea, we use the Gray level residual field (GLR) [35] as the absolute gray-value difference between the actual deformed image and the warped reference image:

$$GLR(X) = \left|I_{def}(X) - I_{pre}(X)\right| \tag{19}$$

where the first and second image terms denote the real deformed image and the reconstructed deformed image, respectively. The latter is obtained by warping the reference image with the estimated displacement field. A lower GLR indicates better gray-field reconstruction consistency and, therefore, a more reliable displacement estimate.

To provide a fair comparison, GLR was evaluated in two regions. The first is the full valid-overlap region, which reflects the overall reconstruction quality after warping. The second is a common central ROI shared by both the proposed method and VIC2D. This common ROI excludes boundary-invalid areas, newly exposed regions, and disappearing speckle regions caused by large rotation. Therefore, the common-ROI GLR mainly evaluates the gray-field consistency in the reliable overlapping speckle region, while the full-region GLR also reflects each method’s robustness to boundary invalidity.

Figure 12 shows the displacement fields and GLR maps for the three real rotation cases. For all tested rotation angles, the proposed method produces smooth and physically consistent displacement fields in both the horizontal and vertical directions. The corresponding GLR maps remain low within the valid speckle region, indicating that the warped reference image is well aligned with the actual deformed image. In contrast, VIC2D produces comparable global displacement trends in some central regions but shows stronger boundary artifacts and higher residual values near invalid or newly exposed regions. This difference becomes more evident as the rotation angle increases.

The quantitative GLR results are summarized in Table 4. In the common ROI, the proposed method achieves lower mean GLR than VIC2D, demonstrating better gray-field reconstruction consistency in the reliable overlapping speckle region. In the full valid region, the advantage of the proposed method is more pronounced because the valid-overlap mask and structural matching loss reduce the influence of invalid boundary regions. These results confirm that the proposed framework can provide stable displacement estimation under real large-rotation and illumination-varying conditions.

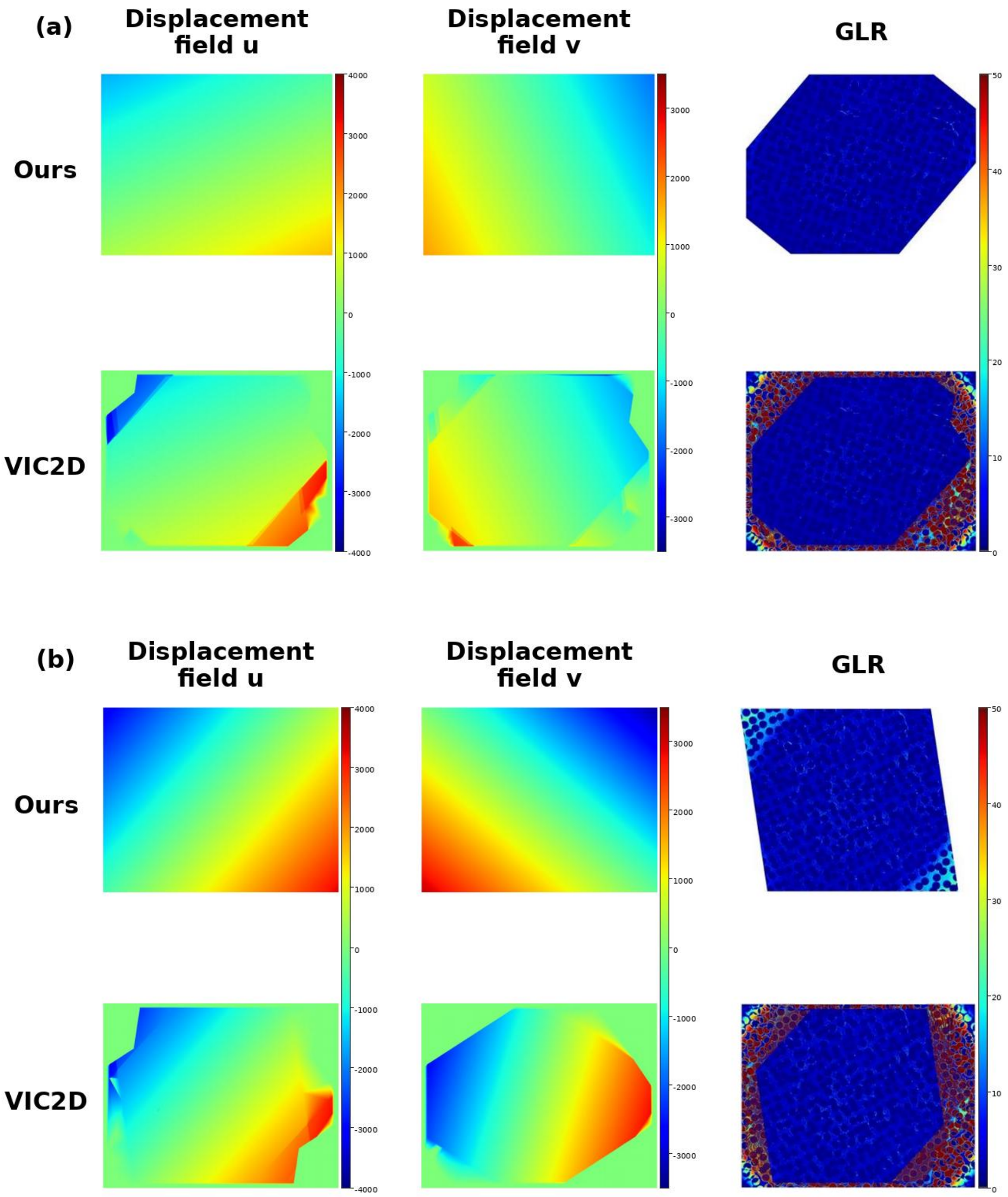

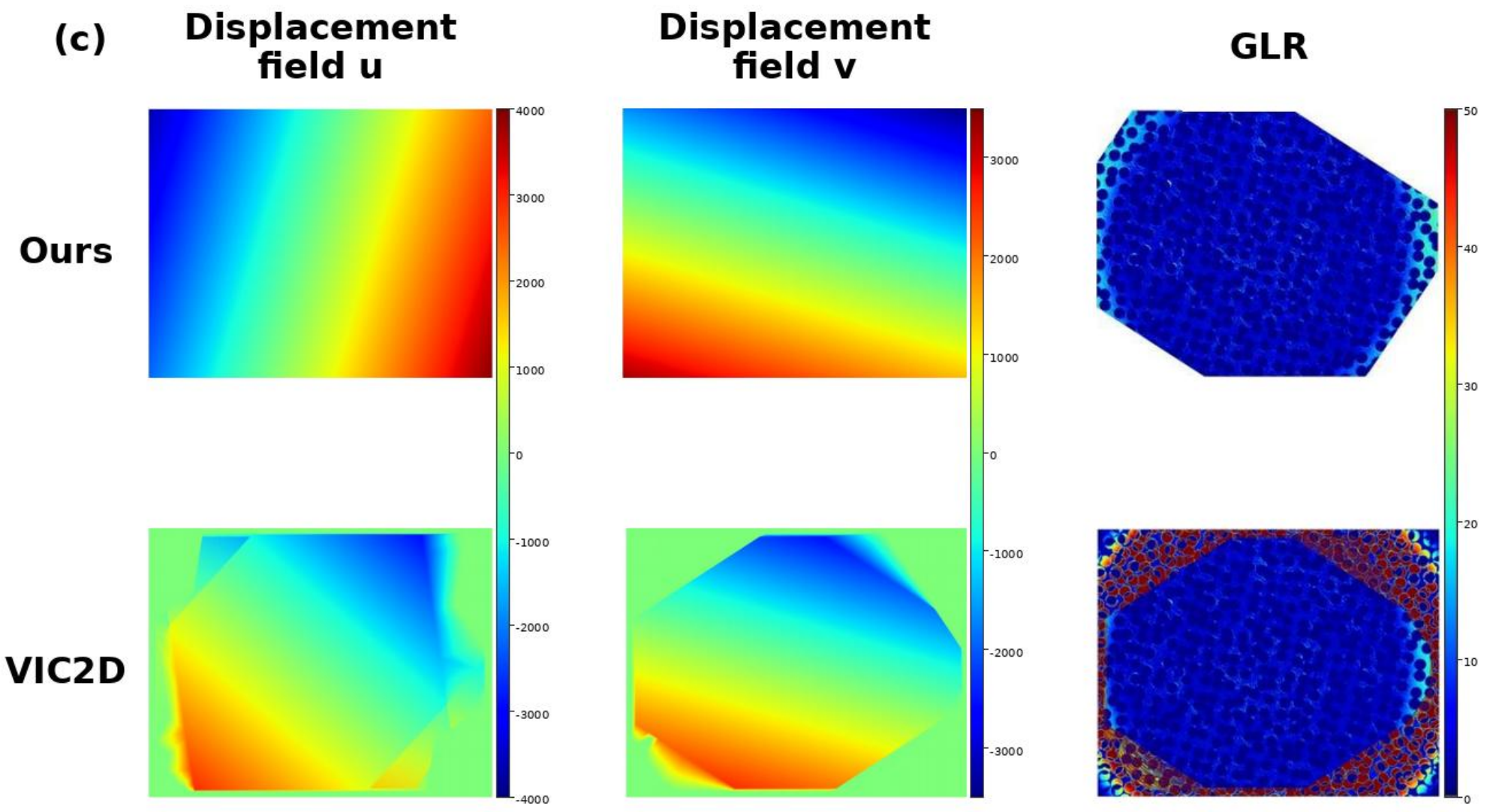

**Fig. 12. Real rotation experimental results and gray-field consistency evaluation.** (a) Approximately 50°rotation. (b) Approximately 100° rotation. (c) Approximately 150° rotation. For each case, the horizontal displacement field $u$, vertical displacement field $v$, and a GLR map are shown for the proposed method and VIC2D. GLR is computed as the absolute gray-value difference between the actual deformed image and the warped reference image. The proposed method produces smoother displacement fields and lower GLR values in the valid speckle region, indicating better gray-field reconstruction consistency under large rotation and illumination variation.

| Rotation angle | Methods | Mean GLR, full valid region | Mean GLR, common ROI |
|---|---|---|---|
| **50°** | VIC2D | 11.92 | 1.68 |
| | **Ours** | **1.67** | **1.65** |
| **100°** | VIC2D | 15.21 | 2.26 |
| | **Ours** | **2.72** | **2.16** |
| **150°** | VIC2D | 14.13 | 2.66 |
| | **Ours** | **3.21** | **2.52** |

**Table 4. Gray-field consistency comparison using GLR in the real rotation experiments.** Mean GLR are computed between the actual deformed image and the warped reference image. The full valid region is determined by the valid-overlap mask after warping, while the common ROI is defined as a $1000 \times 1000$ pixel square region cropped from the image center to reduce the influence of invalid boundary areas and newly exposed regions. Lower values indicate better gray-field reconstruction consistency.

### 3.6. Discussion of experimental results

The simulation and real experimental results demonstrate the effectiveness of the proposed two-stage hybrid neural DIC framework from three aspects. First, in the regular deformation experiments, including noise-contaminated sub-pixel translation, rigid translation, and tensile deformation, the proposed method maintains accurate displacement measurement under standard DIC conditions. These results indicate that the introduction of global alignment and the hybrid KAN-MLP residual representation does not compromise the basic measurement capability of coordinate-based neural DIC. Compared with PINN-DIC and KAN-DIC, the proposed method generally produces smoother displacement fields and lower spatial error fluctuations, especially when image noise or mild intensity disturbance is present.

Second, the large rotation experiments verify the necessity of the Stage-1 masked global rigid alignment. Direct coordinate-based optimization may become unstable when the initial correspondence between the reference and deformed images is poor, because large rotation introduces spatially varying displacement over the whole image. By combining coarse rotation search, phase correlation, and masked loss-based refinement, the proposed method first removes the dominant rigid-body motion before dense residual deformation learning. This global-to-local strategy reduces the optimization difficulty and improves the robustness of neural DIC under large rigid motion.

Third, the degraded image simulations and real rotation experiment show the advantage of the proposed illumination-robust structural matching loss. Raw photometric matching is sensitive to non-uniform illumination, brightness drift, haze-like degradation, and invalid boundary regions caused by warping. In contrast, the local zero-normalized Sobel loss compares normalized structural gradients rather than raw gray values, which improves robustness under practical imaging disturbances. The real experimental results further confirm that the proposed method can obtain a reasonable gray-field reconstruction under large rotation and illumination variation, even when ground-truth displacement fields are unavailable.

Overall, these experiments suggest that the proposed framework is particularly suitable for DIC scenarios where large global motion, local residual deformation, and image intensity disturbance coexist. The method combines the interpretability of coordinate-based optimization with a more robust alignment and matching strategy. Nevertheless, several limitations remain. First, the current method is still optimization-based and therefore slower than feed-forward deep learning models. Second, the performance of the KAN-MLP neural field may depend on hyperparameters such as KAN grid size and network width. Third, although the real experiment is validated through gray-field consistency, more experiments with independently measured ground-truth displacement or strain fields are needed for further quantitative validation. These issues will be investigated in future work.

## 4. Conclusion

This paper proposes a two-stage hybrid KAN-MLP neural DIC framework for robust full-field displacement measurement. The method addresses sensitivity to large global motion, limited recovery of fine residual deformation, and vulnerability to illumination disturbance. It first performs masked global rigid alignment to remove dominant translation and rotation between the reference and deformed images, and then estimates the remaining dense residual field with a hybrid KAN-MLP neural representation. An illumination-robust structural matching loss reduces the influence of brightness drift, haze-like degradation, local contrast variation, and invalid boundary regions.

The main advantage of the proposed method lies in its global-to-local design. Compared with direct coordinate-based optimization, the Stage-1 alignment provides a more reliable initialization under large rigid-body motion. Compared with pure FCNN-based PINN-DIC or pure KAN-based KAN-DIC, the hybrid KAN-MLP representation combines compact nonlinear coordinate modeling with flexible residual refinement. Compared with raw photometric loss, the proposed structural loss focuses on normalized speckle gradient structures, making the optimization more robust under degraded imaging conditions.

Simulation experiments were conducted on benchmark speckle image pairs and additional degraded image cases. The results show that the proposed method maintains accurate displacement estimation under noise-contaminated sub-pixel translation, rigid translation with different displacement amplitudes, and tensile deformation. In large rotation cases, the proposed method achieves more stable displacement recovery than direct coordinate-based neural DIC methods, demonstrating the effectiveness of the masked global rigid alignment stage. In degraded image simulations, the proposed method produces lower error distributions and more reliable displacement fields, validating the robustness of the local zero-normalized Sobel structural loss.

Real speckle image experiments further demonstrate the practical applicability of the proposed method. Under real rotation and illumination variation, the proposed framework successfully reconstructs the deformed image from the reference image and achieves good gray-field consistency, indicating that the estimated displacement field remains physically meaningful even when ground-truth displacement is unavailable.

Overall, the proposed method provides a robust, self-supervised, and coordinate-based neural DIC framework by integrating global rigid alignment, hybrid neural displacement representation, and illumination-insensitive structural matching. Future work will focus on improving computational efficiency, extending the method to more complex real deformation experiments, and incorporating uncertainty estimation or mechanical constraints for more reliable displacement and strain measurement.